%% file: main.tex
\documentclass{article}

\newif\ifpreprint
\preprinttrue

\ifpreprint
  \usepackage[preprint]{neurips_2026}
\else
  \usepackage{neurips_2026}
\fi
\input{math_commands.tex}

\usepackage{hyperref}
\usepackage{url}
\usepackage{booktabs}
\usepackage{amsfonts}
\usepackage{amsmath}
\usepackage{graphicx}
\usepackage{multirow}
\usepackage{subcaption}
\usepackage{xcolor}
\usepackage{pgfplots}
\pgfplotsset{compat=1.18}
\definecolor{C0}{HTML}{1f77b4}
\definecolor{C1}{HTML}{ff7f0e}
\definecolor{C2}{HTML}{2ca02c}
\definecolor{C3}{HTML}{d62728}
\definecolor{C4}{HTML}{9467bd}

\title{Predicting Where Steering Vectors Succeed}

\author{%
  Jayadev Billa\thanks{Unaffiliated researcher; previously at ISI@USC, Yahoo, Nuance, and BBN.} \\
  \texttt{jbilla2004@gmail.com} \\
}

\newcommand{\lap}{\textsc{lap}}
\newcommand{\alin}{A_{\mathrm{lin}}}
\newcommand{\amlp}{A_{\mathrm{mlp}}}
\newcommand{\probegap}{\Delta}

\begin{document}

\maketitle

\begin{abstract}
Steering vectors work for some concepts and layers but fail for others, and practitioners have no way to predict which setting applies before running an intervention. We introduce the \emph{Linear Accessibility Profile} (\lap{}), a per-layer diagnostic that repurposes the logit lens as a predictor of steering vector effectiveness. The key measure, $\alin$, applies the model's unembedding matrix to intermediate hidden states, requiring no training. Across 24 controlled binary concept families on five models (Pythia-2.8B to Llama-8B), peak $\alin$ predicts steering effectiveness at $\rho = +0.86$ to $+0.91$ and layer selection at $\rho = +0.63$ to $+0.92$. A three-regime framework explains when difference-of-means steering works, when nonlinear methods are needed, and when no method can work. An entity-steering demo confirms the prediction end-to-end: steering at the \lap{}-recommended layer redirects completions on Gemma-2-2B and OLMo-2-1B-Instruct, while the middle layer (the standard heuristic) has no effect on either model.
\end{abstract}

\section{Introduction}

Steering vectors add a direction to the residual stream to shift model behavior. They have been applied to refusal \citep{arditi2024refusal}, truthfulness \citep{li2023iti}, and broader behavioral properties \citep{zou2023representation,turner2023activation}. However, effectiveness varies across concepts and layers, and practitioners currently select steering layers by trial and error. No systematic method predicts which setting will succeed.

The logit lens \citep{nostalgebraist2020logit} applies the unembedding matrix to intermediate hidden states to observe how predictions evolve across layers. \citet{belrose2023tuned} addressed the layer norm mismatch with a learned correction (the ``tuned lens''). These methods characterize what the model ``thinks'' at each layer, but none connects this measurement to the success or failure of steering interventions.

We repurpose the logit lens as a \emph{predictor of steering vector effectiveness}. The resulting framework, the Linear Accessibility Profile (\lap{}), measures at each layer whether a concept is accessible through the model's own output projection and whether that accessibility predicts where steering will work. Prior work selects steering layers heuristically, typically targeting middle layers \citep{turner2023activation,templeton2024scaling}. \lap{} operates at a different level: predicting whether a concept is steerable at all. A concept with high peak $\alin$ (e.g., continent, $\alin = 0.68$) steers effectively; one with low peak $\alin$ (e.g., parity, $\alin = 0.02$) does not. We complement the logit lens with a nonlinear upper bound (a residual MLP) to quantify the \emph{probe gap}, and measure perturbation sensitivity ($\lambda$) to identify layers where the representation is unstable.

We validate \lap{} on single-token next-token completion tasks, where the logit lens gives an unambiguous accuracy metric. Extensions to multi-token settings are discussed in Section~\ref{sec:discussion}. We evaluate primarily on Gemma-2-2B \citep{team2024gemma} and replicate on Llama-3.1-8B \citep{dubey2024llama}, Mistral-7B-v0.3, Qwen2.5-7B, and two non-transformers (Mamba-1.4B, RWKV-1.6B). An entity-steering demo validates \lap{} end-to-end: steering London-answer prompts toward ``Paris'' at the \lap{}-recommended layer redirects completions on both Gemma-2-2B and OLMo-2-1B-Instruct, while the middle layer has no effect on either model.

Our contributions: (1) a connection between logit lens measurements and steering vector effectiveness, validated at two levels (layer selection and steerability prediction) across 24 controlled families and five models; (2) a three-regime framework that explains when difference-of-means steering works, when nonlinear methods are needed, and when no method can work; (3) a controlled experimental design using 25 binary concept families that isolates representation geometry from task-structure confounds.

\section{Related Work}

\paragraph{The logit lens and probing.} \citet{nostalgebraist2020logit} introduced the logit lens; \citet{belrose2023tuned} proposed the tuned lens to address the layer norm mismatch; \citet{yomdin2023jump} examined how predictions change at certain layers. We show that the standard logit lens, despite the mismatch, is a strong predictor of steering effectiveness. By using the model's own unembedding (a fixed, untrained projection), we avoid the selectivity concerns that apply to trained probes \citep{belinkov2022probing}. The trade-off is that we measure alignment with one specific linear projection, not general linear decodability.

\paragraph{Linear representations.} \citet{park2024linear} formalize the linear representation hypothesis and connect it to probing and steering. \citet{nanda2023emergent} observe linear representations in Othello-playing models. The hypothesis has been challenged: \citet{csordas2024recurrent} show nonlinear encodings in small models, and \citet{engels2024not} demonstrate multi-dimensional feature manifolds. We do not assume the hypothesis holds universally; \lap{} measures where and to what degree it holds.

\paragraph{Steering and intervention.} \citet{zou2023representation} introduce representation engineering. \citet{turner2023activation} formalize activation addition. \citet{arditi2024refusal} identify a single direction mediating refusal. Each method demonstrates success on its target concept but does not predict when steering will succeed on a new concept or layer.

\paragraph{Sparse autoencoders and transcoders.} \citet{templeton2024scaling} scale SAEs to large models; \citet{lieberum2024gemma} release GemmaScope transcoders; \citet{ameisen2025circuit} introduce attribution graphs. Our three-regime framework predicts that SAE features should be most useful in regime~2 (concept present but not output-aligned), where difference-of-means fails.

\section{Method}
\label{sec:method}

\subsection{Setup}

Consider a causal language model with $L$ transformer blocks. Each block reads from and writes to a shared \emph{residual stream}: $h_\ell = h_{\ell-1} + \mathrm{block}_\ell(h_{\ell-1})$, where $h_0$ is the token embedding. After the final block, an output head\footnote{All transformer models in this paper use RMSNorm rather than standard LayerNorm. We write $\text{LayerNorm}$ throughout for readability; the distinction does not affect the logit lens analysis.} produces logits: $\mathrm{logits} = W_U \cdot \text{LayerNorm}(h_L)$, where $W_U \in \mathbb{R}^{V \times d}$ is the unembedding matrix. Because the residual stream lives in $\mathbb{R}^d$ at every layer, this output head can be applied to any intermediate $h_\ell$, which is the basis of the logit lens.

For a concept family $\mathcal{C} = \{(x_i, t_i)\}_{i=1}^N$, where each prompt $x_i$ has a correct next-token answer $t_i$, we measure how linearly accessible the concept is at each layer.

\subsection{Linear accuracy (logit lens)}

We apply the model's unembedding to intermediate hidden states:
\begin{equation}
\alin(\ell) = \frac{1}{N} \sum_{i=1}^{N} \begin{cases} 1 & \text{if } \arg\max_v (W_U \cdot \text{LayerNorm}(h_\ell^{(i)}))_v = t_i \\ 0 & \text{otherwise} \end{cases}
\end{equation}
This is the logit lens evaluated as classification accuracy over the concept family. No training is required. We apply the \emph{final} layer norm to intermediate states, inheriting the layer norm mismatch discussed by \citet{belrose2023tuned}. We evaluate the effect of this mismatch in Section~\ref{sec:discussion}.

\subsection{Probe gap}

The logit lens measures what is linearly accessible through the model's output projection, but concept information may be present in a form that requires nonlinear transformation before it aligns with the unembedding. The \emph{probe gap} $\probegap(\ell) = \amlp(\ell) - \alin(\ell)$ quantifies how much concept information is present at layer $\ell$ but not output-aligned.

We train a residual MLP to compute $\amlp$:
\begin{equation}
\hat{h}_\ell = h_\ell + f_\theta(h_\ell), \qquad \amlp(\ell) = \frac{1}{N} \sum_{i=1}^N \begin{cases} 1 & \text{if } \arg\max_v (W_U \cdot \text{LayerNorm}(\hat{h}_\ell^{(i)}))_v = t_i \\ 0 & \text{otherwise} \end{cases}
\end{equation}
where $f_\theta$ is a two-layer MLP ($d \to 512 \to d$) with layer normalization, GELU, and dropout ($p = 0.1$), trained on 80\% of prompts to minimize cross-entropy. The residual connection ensures the MLP learns a correction rather than replacing the hidden state. A large probe gap indicates nonlinear encoding at that layer; steering is unlikely to work even though the information is present.

\subsection{Perturbation sensitivity}

We measure how much a small random perturbation at layer $\ell$ is amplified by subsequent computation:
\begin{equation}
\lambda(\ell) = \frac{1}{K} \sum_{k=1}^{K} \frac{\|f(h_\ell + \alpha \epsilon_k) - f(h_\ell)\|}{\alpha}, \qquad \alpha = 0.01 \cdot \|h_\ell\|
\end{equation}
where $\epsilon_k$ are random unit vectors, $f$ is the forward pass from layer $\ell$ to output logits, and $K = 10$. High $\lambda$ indicates an unstable representation where steering vectors will have unpredictable effects.

\subsection{The Linear Accessibility Profile}

For a concept family $\mathcal{C}$ and layer $\ell$, the \emph{Linear Accessibility Profile} (\lap{}) is:
\begin{equation}
\lap(\ell) = \bigl(\alin(\ell),\; \probegap(\ell),\; \lambda(\ell)\bigr)
\end{equation}
Of these, $\alin$ is the primary predictor. The remaining components characterize why steering may fail: high $\probegap$ means information is present but not output-aligned; high $\lambda$ means the representation is unstable (Figure~\ref{fig:chaotic} in Appendix~\ref{app:chaotic}).

\subsection{Concept families}

We use two sets of concept families. All correct answers are single tokens in the model vocabulary (required because the logit lens produces a distribution over individual tokens).

\paragraph{Core families (5).} Five heterogeneous families (Table~\ref{tab:families}) are used for \emph{within-concept} analyses: measuring how $\alin$, $\probegap$, and $\lambda$ vary across layers for a single concept.

\begin{table}[t]
\centering
\caption{Core concept families (used for within-concept layer analysis). All correct answers are single tokens.}
\label{tab:families}
\small
\begin{tabular}{llrl}
\toprule
Family & Example prompt & $N$ & Correct token \\
\midrule
Arithmetic & ``2 + 5 = '' & 831 & ``7'' \\
Geography & ``Paris is the capital of'' & 564 & ``France'' \\
Sequence & ``Monday, Tuesday, Wednesday,'' & 433 & ``Thursday'' \\
Word transform & ``The opposite of hot is'' & 506 & ``cold'' \\
Analogy & ``hot is to cold as big is to'' & 348 & ``small'' \\
\bottomrule
\end{tabular}
\end{table}

\paragraph{Controlled binary families (25).} For \emph{steerability prediction} across concepts, task-structure confounds must be eliminated. The core families vary in answer-class count, target sizes, and prompt formats; comparing steerability across them yields a non-significant correlation ($\rho = +0.18$, $p = 0.54$). We construct 25 controlled binary families (Table~\ref{tab:controlled} in the appendix): each has two answer classes, balanced groups ($\sim$22 items per class), and consistent templates. This reveals the underlying signal ($\rho = +0.86$ to $+0.91$, $p < 10^{-3}$; details in Appendix~\ref{app:controlled_method}).

\section{Experiments}

We evaluate primarily on Gemma-2-2B (26 layers, $d = 2304$) with replication on Llama-3.1-8B (32 layers), Mistral-7B-v0.3 (32 layers), Qwen2.5-7B (28 layers), and two non-transformer architectures: Mamba-1.4B (48 layers) and RWKV-1.6B (24 layers). Table~\ref{tab:models} in the appendix specifies which model is used for each experiment.

\subsection{Linear accessibility across layers}
\label{sec:exp1}

Table~\ref{tab:exp1} reports the main results. All five families show zero linear accuracy for layers 0--15 and sharp emergence in layers 18--24, consistent with the logit lens literature \citep{nostalgebraist2020logit,yomdin2023jump}. Accuracy peaks at layer 23--24 (not the final layer) for four of five families.

\begin{table}[t]
\centering
\caption{Linear accessibility across layers of Gemma-2-2B. $\alin$: logit lens accuracy at the best layer. $\amlp$: MLP probe accuracy. $\Delta$: probe gap. Acc(a): $\alin$ on prompts the model answers correctly. Acc(b): $\alin$ on prompts the model answers incorrectly.}
\label{tab:exp1}
\small
\begin{tabular}{lcccccc}
\toprule
Family & Model top-1 & Best $\alin$ & Best $\amlp$ & $\Delta$ & Acc(a) & Acc(b) \\
\midrule
Sequence & 77.8\% & 0.711 (L24) & 0.954 & 0.243 & 0.914 & 0.000 \\
Arithmetic & 65.7\% & 0.686 (L23) & 0.904 & 0.218 & 0.954 & 0.172 \\
Word transform & 62.3\% & 0.512 (L24) & 0.647 & 0.135 & 0.762 & 0.099 \\
Analogy & 71.6\% & 0.451 (L24) & 0.629 & 0.177 & 0.621 & 0.030 \\
Geography & 60.1\% & 0.280 (L25) & 1.000 & 0.720 & 0.466 & 0.009 \\
\bottomrule
\end{tabular}
\end{table}

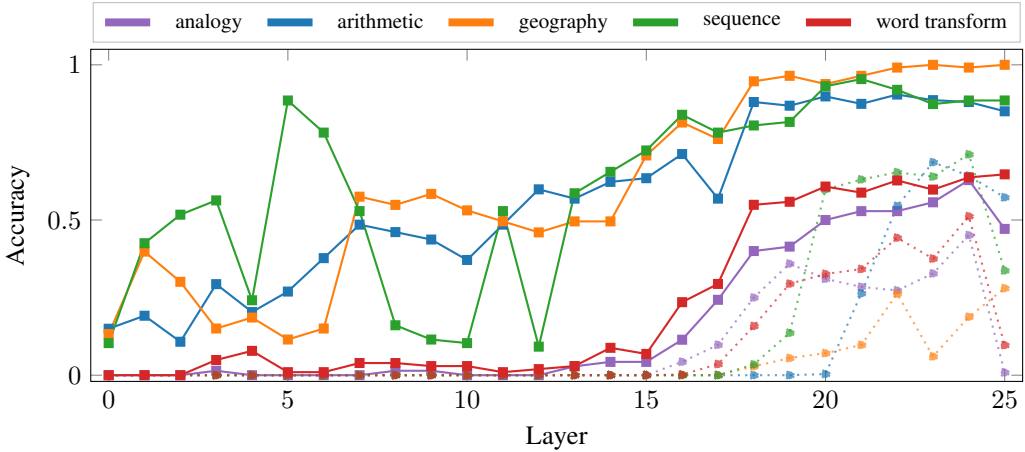
\begin{figure}[t]
\centering
\begin{tikzpicture}
\begin{axis}[
    width=\textwidth, height=6cm,
    xlabel={Layer}, ylabel={Accuracy},
    xmin=-0.5, xmax=25.5, ymin=-0.02, ymax=1.05,
    xtick={0,5,10,15,20,25},
    legend columns=-1,
    legend style={at={(0.5,1.02)}, anchor=south, font=\footnotesize, draw=gray!50,
                  column sep=6pt},
    every axis plot/.append style={thick},
]
  \addplot[C4, solid, mark=square*, mark size=1.5, forget plot] table[x=layer, y=analogyamlp, col sep=comma] {data/fig1_emergence.csv};
  \addplot[C4, dotted, mark=*, mark size=1.5, opacity=0.7, forget plot] table[x=layer, y=analogyalin, col sep=comma] {data/fig1_emergence.csv};
  \addplot[C0, solid, mark=square*, mark size=1.5, forget plot] table[x=layer, y=arithmeticamlp, col sep=comma] {data/fig1_emergence.csv};
  \addplot[C0, dotted, mark=*, mark size=1.5, opacity=0.7, forget plot] table[x=layer, y=arithmeticalin, col sep=comma] {data/fig1_emergence.csv};
  \addplot[C1, solid, mark=square*, mark size=1.5, forget plot] table[x=layer, y=geographyamlp, col sep=comma] {data/fig1_emergence.csv};
  \addplot[C1, dotted, mark=*, mark size=1.5, opacity=0.7, forget plot] table[x=layer, y=geographyalin, col sep=comma] {data/fig1_emergence.csv};
  \addplot[C2, solid, mark=square*, mark size=1.5, forget plot] table[x=layer, y=sequenceamlp, col sep=comma] {data/fig1_emergence.csv};
  \addplot[C2, dotted, mark=*, mark size=1.5, opacity=0.7, forget plot] table[x=layer, y=sequencealin, col sep=comma] {data/fig1_emergence.csv};
  \addplot[C3, solid, mark=square*, mark size=1.5, forget plot] table[x=layer, y=wordtransformamlp, col sep=comma] {data/fig1_emergence.csv};
  \addplot[C3, dotted, mark=*, mark size=1.5, opacity=0.7, forget plot] table[x=layer, y=wordtransformalin, col sep=comma] {data/fig1_emergence.csv};

  \addplot[C4, line width=3pt, no markers] coordinates {(0,0)};  \addlegendentry{analogy}
  \addplot[C0, line width=3pt, no markers] coordinates {(0,0)};  \addlegendentry{arithmetic}
  \addplot[C1, line width=3pt, no markers] coordinates {(0,0)};  \addlegendentry{geography}
  \addplot[C2, line width=3pt, no markers] coordinates {(0,0)};  \addlegendentry{sequence}
  \addplot[C3, line width=3pt, no markers] coordinates {(0,0)};  \addlegendentry{word transform}
\end{axis}
\end{tikzpicture}
\caption{Per-layer $\amlp$ (solid) and $\alin$ (dotted) for each concept family on Gemma-2-2B. The gap between solid and dotted lines is the probe gap $\probegap$. All families show $\alin = 0$ at layers 0--15 and sharp emergence in layers 18--24. The nonlinear probe detects concepts substantially earlier---sequence reaches $\amlp > 0.9$ at layer~5, while $\alin$ remains zero until layer~18.}
\label{fig:emergence}
\end{figure}

The probe gap varies widely. For arithmetic and sequence, $\Delta \approx 0.22$: the concept is predominantly linear at the best layer. For geography, $\Delta = 0.720$: the MLP achieves perfect accuracy while the logit lens reaches only 28.0\%. The MLP also detects concepts earlier: sequence is nonlinearly accessible at layer~5 ($\amlp = 0.91$) but not linearly until layer~20 ($\alin = 0.60$).

\paragraph{Crystallization gap.} We define the gap between nonlinear detection ($\amlp > 0.5$) and linear emergence ($\alin > 0.1$) as the \emph{crystallization gap}. Both metrics measure argmax accuracy over the full vocabulary ($\sim$256K tokens), so chance is effectively zero. The $\amlp$ threshold of 50\% indicates the nonlinear probe recovers the correct token for a majority of prompts. The $\alin$ threshold is lower at 10\% because sporadic values of 1--3\% can arise from token frequency biases in the unembedding; 10\% requires a substantial fraction of prompts to have the correct token as the logit lens argmax. The gap varies systematically (Table~\ref{tab:crystal} in the appendix): sequence is nonlinearly detectable at layer~1 but not linearly accessible until layer~19, a gap of 18 layers (69\% of depth). Word transformation and analogy show the opposite pattern, with $\alin$ emerging before $\amlp$ reaches 0.5, suggesting these concepts are natively aligned with the unembedding. Tuned lens comparisons shift emergence 2--5 layers earlier but do not eliminate the gap, confirming that the crystallization gap reflects genuine nonlinear encoding rather than layer norm mismatch alone.

We also observe that $\alin$ tracks the model's own accuracy. Table~\ref{tab:exp1} splits $\alin$ by whether the model answers each prompt correctly (Acc(a)) or incorrectly (Acc(b)). On prompts where the model's top-1 prediction is correct, $\alin$ ranges from 0.47 to 0.95; on prompts where the model gets the answer wrong, $\alin$ is 0.00 to 0.17. Linear accessibility appears to be a necessary condition for correct output.

\subsection{Predicting steering vector effectiveness}
\label{sec:exp2}

For each concept family, we select a target answer (one specific correct token that appears frequently), split prompts into target and non-target groups, compute the steering direction as $d_\ell = \bar{h}_\ell^{\,\text{target}} - \bar{h}_\ell^{\,\text{non-target}}$ at each layer, inject $d_\ell$ into non-target prompts, and measure the change in target-token probability ($\Delta P$).

\paragraph{Raw correlations.} Table~\ref{tab:exp2} reports Spearman $\rho$ between $\alin$ and $\Delta P$ per family. Correlations range from $+0.72$ to $+0.87$, all $p < 10^{-3}$.

\begin{table}[t]
\centering
\caption{Per-family steering correlations on Gemma-2-2B (26 layers). $\rho(\alin, \Delta P)$: Spearman correlation between logit lens accuracy and steering effect across layers. $\rho(\lambda, \Delta P)$: perturbation sensitivity vs.\ steering. Partial $r$: Pearson partial correlation controlling for layer index. All correlations are across the 26 layers of Gemma-2-2B.}
\label{tab:exp2}
\small
\begin{tabular}{lcccc}
\toprule
Family & $\rho(\alin, \Delta P)$ & $\rho(\lambda, \Delta P)$ & Partial $r(\alin, \Delta P \mid \ell)$ & $p$ \\
\midrule
Word transform & $+0.871$ & $-0.744$ & $+0.809$ & $< 10^{-4}$ \\
Sequence & $+0.814$ & $-0.830$ & $+0.869$ & $< 10^{-4}$ \\
Arithmetic & $+0.719$ & $-0.990$ & $+0.688$ & $< 10^{-3}$ \\
Analogy & $+0.866$ & $-0.923$ & $+0.647$ & $0.001$ \\
Geography & $+0.758$ & $-0.729$ & $+0.372$ & $0.067$ \\
\midrule
Pooled ($n = 130$) & $+0.777$ & $-0.706$ & $+0.507$ & $< 10^{-9}$ \\
\bottomrule
\end{tabular}
\end{table}

\paragraph{Controlling for depth.} Both $\alin$ and $\Delta P$ increase with layer depth. The pooled partial correlation controlling for layer index is $r = +0.507$ ($p < 10^{-9}$, $n = 130$); four of five families remain individually significant (Table~\ref{tab:exp2}). Details of the restricted analysis and permutation test are in Appendix~\ref{app:depth}.

\paragraph{Steerability prediction.} The more practically important question is whether a concept is steerable at all, and by what method. The full LAP profile identifies three regimes. When $\amlp$ is low, the concept is not yet represented in the residual stream and no steering method can work, regardless of technique. When $\amlp$ is high but $\alin$ is low (the crystallization gap), the concept is present but nonlinearly encoded; difference-of-means steering fails because the separating direction does not align with the output projection, though nonlinear methods such as SAE feature amplification \citep{templeton2024scaling} could potentially extract the concept. When $\alin$ is high, the concept is output-aligned and difference-of-means steering works. $\alin$ thus serves as a diagnostic for the choice of steering method: it tells the practitioner whether the simpler difference-of-means approach will suffice, or whether a more sophisticated technique is needed.

The standard heuristic is to steer at the middle layer \citep{turner2023activation,templeton2024scaling}. The three regimes explain both its successes and failures. In our refusal demo on Llama-3.2-1B-Instruct (Appendix~\ref{app:demos}), the refusal direction has separability 0.988 at the middle layer, so difference-of-means works. For factual concepts on Gemma-2-2B, the middle layer is in regime~2: all five families show $\amlp > 0.7$ at L15 but $\alin = 0$ (Figure~\ref{fig:emergence}). The concepts are present but not output-aligned, and steering has no effect. Steering works only at late layers (L22 on Gemma, L13 on OLMo) where $\alin > 0$. The Golden Gate Claude result \citep{templeton2024scaling} is consistent with regime~2: SAE feature amplification can extract concept information that is present nonlinearly.

$\alin$ provides a concept-specific prediction. Continent has peak $\alin = 0.70$ and steers effectively ($\Delta P = +0.77$ on Llama-8B); parity-related concepts with peak $\alin < 0.05$ do not steer (typical $\Delta P < 0.01$). To test this cleanly, we use the 25 controlled binary families described in Section~3.6.

\begin{figure}[t]
\centering
\begin{tikzpicture}
\begin{axis}[
    name=left,
    width=0.5\textwidth, height=5.5cm,
    xlabel={Peak $A_{\mathrm{lin}}$},
    ylabel={Max steering $\Delta P$},
    title={\small Gemma-2-2B ($\rho = +0.86$)},
    xmin=-0.02, xmax=0.75, ymin=-0.02, ymax=0.25,
    every axis plot/.append style={only marks, mark=*, mark size=2pt, C0},
    clip=false,
]
\addplot table[x=peakalin, y=maxdp, col sep=comma] {data/fig2_gemma22b.csv};
\end{axis}
\begin{axis}[
    at={(left.east)}, anchor=west, xshift=0.8cm,
    width=0.5\textwidth, height=5.5cm,
    xlabel={Peak $A_{\mathrm{lin}}$},
    ylabel={},
    title={\small Qwen-7B ($\rho = +0.90$)},
    xmin=-0.02, xmax=0.75, ymin=-0.02, ymax=0.30,
    every axis plot/.append style={only marks, mark=*, mark size=2pt, C3},
    clip=false,
]
\addplot table[x=peakalin, y=maxdp, col sep=comma] {data/fig2_Qwen257B.csv};
\end{axis}
\end{tikzpicture}
\caption{Steerability prediction: peak $\alin$ vs.\ max steering $\Delta P$ for 24 controlled binary concept families. Each point is one concept family. Higher $\alin$ predicts stronger steering. The correlation is $\rho = +0.86$ on Gemma-2-2B and $\rho = +0.90$ on Qwen-7B, replicated across five models (Table~\ref{tab:crossconcept}).}
\label{fig:crossconcept}
\end{figure}
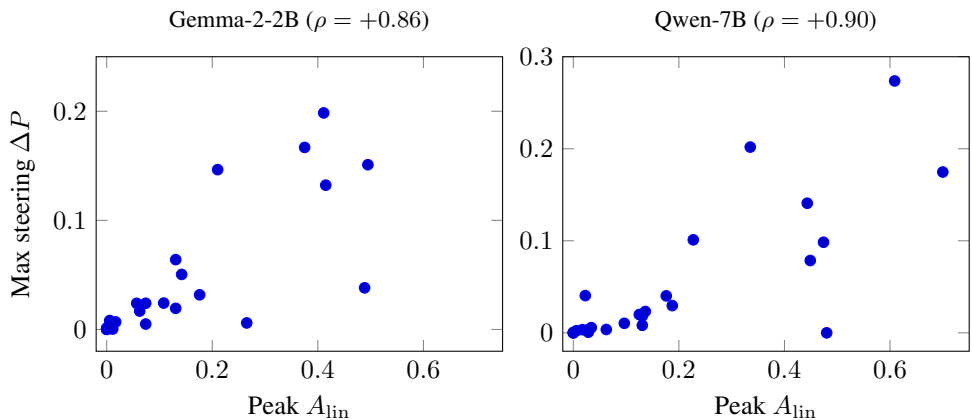

Across these controlled families (Figure~\ref{fig:crossconcept}), peak $\alin$ predicts maximum steering $\Delta P$ with $\rho = +0.86$ to $+0.91$ across five models (all $p < 10^{-3}$). The correlation strengthens with model size and holds above the $\alin = 0$ floor on larger models: restricting to families with $\alin > 0.1$, $\rho = +0.86$ on Qwen-7B ($n = 13$) and $\rho = +0.84$ on Llama-8B ($n = 14$).

\paragraph{The output-alignment principle.} Why does $\alin$ predict steering? Steering adds a direction $d$ to $h_\ell$, and the model's readout is approximately linear: $C(h_\ell + d) \approx C(h_\ell) + C(d)$, where $C = W_U \circ \text{LayerNorm}$. The steering effect is governed by $C(d)$: the steering direction projected through the unembedding. If $C(d)$ assigns high weight to the target token, steering shifts probability toward it; if $C(d)$ points at unrelated tokens, the perturbation produces unpredictable changes. $\alin$ measures exactly this alignment. Empirically, $C(d)_{\mathrm{target}}$ correlates with $\Delta P$ ($\rho = +0.73$ to $+0.84$ across three models, all $p < 10^{-4}$), and $\alin$ is a stronger predictor ($\rho = +0.86$ to $+0.90$) because it captures the concept's overall organization in the output space.

\paragraph{Collateral damage and steering efficiency.} We define \emph{steering efficiency} as $\Delta P / \mathrm{KL}_{\mathrm{collateral}}$, where $\mathrm{KL}_{\mathrm{collateral}}$ measures the KL divergence on 50 unrelated prompts. On Gemma-2-2B, steering at the $\alin$-recommended layer produces substantially higher efficiency than the middle layer for all five families: sequence is $+1.31$ at L24 vs.\ $+0.000$ at L13; arithmetic is $+0.84$ vs.\ $+0.094$ (full comparison in Table~\ref{tab:efficiency}; analysis in Appendix~\ref{app:theory}). Across concepts, peak $\alin$ correlates with efficiency at $\rho = +0.63$ to $+0.65$ across three models (Gemma-2-2B, Qwen-1.5B, Pythia-2.8B; all $p < 10^{-3}$).

\paragraph{Cross-model consistency.} Concepts that are linearly separable in one model's output space should be linearly separable in another trained on similar data. We observe this: $\alin$ on Gemma predicts $\Delta P$ on Qwen-7B ($\rho = +0.85$) and Llama-8B ($\rho = +0.88$), nearly as strong as same-model prediction. A practitioner can screen concepts on a cheap model: $\alin$ on Qwen-1.5B predicts $\Delta P$ on Llama-8B ($\rho = +0.85$) and Qwen-7B ($\rho = +0.88$).

\paragraph{Scaling with model size.} Across five Pythia models (160M to 6.9B), mean peak $\alin$ increases with scale (0.055 to 0.118) and the number of steerable concepts grows (5/23 to 9/23). Mean steering $\Delta P$ increases roughly $3.4\times$ (0.011 to 0.037). Both $\alin$ and the steerable count plateau between 2.8B and 6.9B. The cross-family correlation $\rho(\alin, \Delta P)$ strengthens with scale ($+0.42$ at 160M to $+0.86$ at 6.9B; Table~\ref{tab:scaling}), suggesting the diagnostic becomes more informative on larger models.

\paragraph{Layer selection and trained probe failure.} Steering at the highest-$\alin$ layer matches or near-matches the oracle for 4 of 5 families (Table~\ref{tab:decision} in the appendix). A trained logistic regression probe achieves $>$93\% classification accuracy at every layer from L0 to L25, yet steering at early layers produces zero $\Delta P$. What matters for steering is not whether some linear separator exists in the high-dimensional residual stream, but whether the concept is aligned with the model's own output projection.

\subsection{Generalization}
\label{sec:exp5}

\paragraph{Cross-architecture replication.} We replicate on Llama-3.1-8B, Mistral-7B-v0.3, and Qwen2.5-7B (Table~\ref{tab:replication_alin}). The emergence pattern holds across all four architectures: zero in the first 70--80\% of depth, sharp rise in the final quarter. The steering correlation is positive across all 12 model--family pairs ($\rho$ range $+0.66$ to $+0.93$). Mistral required a tokenizer-specific fix (Appendix~\ref{app:mistral}).

\begin{table}[t]
\centering
\caption{Peak linear accessibility ($\alin$) and steering correlation ($\rho$) across four architectures. ``---'': steering targets too small ($n < 10$). Mistral required a tokenizer-specific fix for digits (see text).}
\label{tab:replication_alin}
\small
\begin{tabular}{lcccc}
\toprule
& Gemma-2-2B & Llama-8B & Mistral-7B & Qwen-7B \\
& (26 layers) & (32 layers) & (32 layers) & (28 layers) \\
\midrule
\multicolumn{5}{l}{\emph{Peak $\alin$ (layer)}} \\
\addlinespace
Arithmetic & 0.686 (L23) & 0.995 (L29) & 0.695 (L30) & 0.935 (L26) \\
Geography & 0.284 (L25) & 0.680 (L25) & 0.540 (L25) & 0.590 (L25) \\
Sequence & 0.711 (L24) & 0.820 (L31) & 0.735 (L31) & 0.780 (L26) \\
Word transform & 0.512 (L24) & 0.715 (L31) & 0.650 (L31) & 0.500 (L26) \\
Analogy & 0.451 (L24) & 0.415 (L29) & 0.765 (L31) & 0.385 (L26) \\
\midrule
\multicolumn{5}{l}{\emph{Steering $\rho(\alin, \Delta P)$}} \\
\addlinespace
Arithmetic & $+0.719$ & $+0.849$ & $+0.886$ & $+0.707$ \\
Geography & $+0.758$ & $+0.930$ & $+0.785$ & $+0.717$ \\
Sequence & $+0.814$ & $+0.904$ & $+0.885$ & $+0.657$ \\
\bottomrule
\end{tabular}
\end{table}

\paragraph{Non-transformer architectures.} We replicate the logit lens on Mamba-1.4B (48-layer SSM; \citealt{gu2024mamba}) and RWKV-1.6B (24-layer linear attention; \citealt{peng2023rwkv}). Both exhibit the same emergence pattern and final-layer drop consistent with the layer norm mismatch. Peak $\alin$ closely tracks model accuracy (Table~\ref{tab:replication_ssm} in the appendix). Steering experiments on non-transformers are left for future work.

\paragraph{Entity steering demo.} To validate \lap{} end-to-end, we steer London-answer prompts toward ``Paris'' on two architectures. On Gemma-2-2B, steering at L22 ($\alin = 0.20$) redirects completions (``Big Ben is located in'' $\to$ ``the heart of Paris, France''), while the middle layer L13 ($\alin = 0$) has no effect ($\rho(\alin, \Delta P) = +0.663$, $p < 0.001$). On OLMo-2-1B-Instruct, steering at L13 ($\alin = 0.60$) produces clean redirection (``The capital of England is'' $\to$ ``Paris''), while L8 ($\alin = 0$) again fails ($\rho = +0.753$, $p < 0.001$). This is a direct comparison against the practitioner heuristic: the standard ``steer at the middle layer'' approach fails on both models, while \lap{} identifies the correct layer. Full generation examples are in Appendix~\ref{app:demos}.

\section{Discussion}
\label{sec:discussion}

\paragraph{Tuned lens comparison.} We trained tuned lenses on Pythia-2.8B, Qwen-1.5B, and Gemma-2-2B using the Muon optimizer \citep{jordan2024muon} on 100M tokens. The ``final-layer anomaly'' ($\alin$ dropping to zero at the last layer) disappears with the tuned lens on Qwen and Pythia, confirming it is a layer norm mismatch artifact on those architectures. Gemma's final layer behaves differently: the tuned lens does not resolve the drop, suggesting a genuinely different representation structure. For steering layer selection, the raw logit lens is the better predictor (it measures exactly what the steering mechanism uses), while the tuned lens is more informative for understanding when concepts first emerge. Full per-family comparisons are in Table~\ref{tab:lens_comparison} (Appendix~\ref{app:tunedlens}).

\paragraph{Connection to frontier-scale interpretability.} Concurrent work by \citet{sofroniew2026emotions} validates linear steering at frontier scale: 171 emotion concepts in Claude Sonnet~4.5 are representable as linear directions with causal behavioral effects. Their analysis projects steering directions through the unembedding, and finds that emotion representations concentrate in middle-to-late layers, consistent with our emergence pattern.

\paragraph{SAE prediction.} Our three-regime framework predicts that SAE features should be most useful for steering in regime~2. In regime~3, where the concept is already linearly accessible, SAE decomposition should add less value over difference-of-means. Testing this prediction is a natural direction for future work.

\paragraph{When $\alin$ fails.} Two families have moderate $\alin$ but low $\Delta P$. Parity ($\alin = 0.27$, $\Delta P = +0.003$): the $C(d)$ target rank is 13,440, meaning the separation direction does not align with the ``odd'' token. Gender ($\alin = 0.50$, $\Delta P = +0.012$): $C(d)$ correctly points at ``she,'' but baseline probability is already $\sim$0.4, leaving little room for improvement. These cases suggest $\alin$ should be combined with a baseline-probability check.

\paragraph{Scope and limitations.} Practical guidelines for applying \lap{}, including go/no-go thresholds and layer-selection rules, are in Appendix~\ref{app:guidelines}. The steerability correlation includes a floor effect: concepts with $\alin \approx 0$ trivially have $\Delta P \approx 0$. The above-floor correlation ($\alin > 0.1$) is moderate on smaller models ($\rho = +0.50$ to $+0.52$ on Pythia-2.8B and Gemma-2-2B) and strong on larger models ($\rho = +0.84$ to $+0.93$ on Qwen-7B, Llama-8B, and Qwen-1.5B). We validated on single-token tasks because they provide the cleanest measurement setting; extensions to multi-token and distribution-level settings are discussed in Appendix~\ref{app:multitoken}.

\section{Conclusion}

The logit lens, a training-free measurement, predicts where steering vectors succeed. It predicts which layer to steer at ($\rho = +0.63$ to $+0.92$) and which concepts are steerable ($\rho = +0.86$ to $+0.91$ across five models, with above-floor $\rho = +0.86$ on Qwen-7B). The mechanism is straightforward: the steering direction, projected through the unembedding, must assign high weight to the target token, and $\alin$ measures exactly this. We provide a practitioner-facing diagnostic: compute $\alin$ in one forward pass to determine whether steering is viable, at which layer, and how clean the effect will be.

\paragraph{Reproducibility.} All experiments use publicly available models (Gemma-2-2B, Pythia-160M/410M/1B/2.8B/6.9B, Qwen2.5-1.5B/7B, Llama-3.1-8B, Llama-3.2-1B/Instruct, Mistral-7B, OLMo-2-1B-Instruct, Mamba-1.4B, RWKV-1.6B). Experiments were conducted on a single NVIDIA RTX 3090 Ti GPU (24\,GB). Total compute: primary experiments $\sim$4 GPU-hours; steerability analysis $\sim$2 GPU-hours per model; tuned lens training $\sim$8 GPU-hours per model. The diagnostic tool, experimental code, and 25 controlled concept families are provided as anonymized supplementary material.

\begin{ack}
The author used Claude (Anthropic) and Claude Code during preparation for manuscript critique, narrative feedback, literature search, and experiment implementation and debugging. All research design, theoretical development, experimental execution, analysis, and writing are the author's own. The author takes full responsibility for all content.

The code repository backing this paper is available at \url{https://github.com/jb1999/lap-steering-paper}. No external funding supported this work; the author declares no competing interests.
\end{ack}

\bibliographystyle{plainnat}
\bibliography{references}

\appendix

\section{Practical guidelines}
\label{app:guidelines}

A practitioner wanting to steer a model on a new concept faces two questions: (a) will steering work for this concept? and (b) at which layer? \lap{} addresses both:

\begin{enumerate}
\item \textbf{Compute $\alin(\ell)$.} One forward pass per prompt through the frozen model, applying the unembedding to each layer's hidden state. No training required. Cost: $L$ matrix multiplications per prompt.
\item \textbf{Will steering work? (go/no-go)} If peak $\alin < 0.05$ across all layers, the concept is not linearly accessible and difference-of-means steering will produce negligible effects. If peak $\alin > 0.1$, steering is likely viable. This prediction holds across 24 controlled concept families on five models ($\rho = +0.86$ to $+0.91$ between peak $\alin$ and max $\Delta P$).
\item \textbf{Which layer?} Steer at the layer with the highest $\alin$. This matches or near-matches the oracle-best steering layer for 4 of 5 families on Gemma-2-2B ($\geq$99.7\% of oracle $\Delta P$), with within-family correlations of $\rho = +0.63$ to $+0.92$. The one miss (geography) has the weakest $\alin$ signal (peak 0.28) and the largest probe gap.
\item \textbf{Expect clean steering when $\alin$ is high.} Concepts with high $\alin$ achieve more on-target effect per unit of collateral damage ($\rho = +0.63$ to $+0.65$ across three models). For deployment-sensitive applications, prefer steering at layers with high $\alin$.
\item \textbf{Screen on a small model.} $\alin$ on one model predicts steerability on another ($\rho = +0.64$ to $+0.88$ cross-model). Compute $\alin$ on a small, fast model to identify promising concepts before investing compute on the target model.
\item \textbf{What if steering is weak despite high $\alin$?} Possible causes: (a) the steering target has too few prompts for a clean direction ($n_{\mathrm{target}} < 10$); (b) the concept has many answer classes, diluting the binary contrast; (c) the model's knowledge is split across near-synonym tokens. Examine the model's output distribution at the recommended layer.
\end{enumerate}

These thresholds are calibrated on Gemma-2-2B, Qwen-1.5B, and Pythia-2.8B and validated on Qwen-7B and Llama-8B. The emergence pattern and layer ordering are robust across models; absolute thresholds may need adjustment for larger models.

\paragraph{Compute cost.} \lap{} requires one forward pass plus $L$ unembedding multiplications ($\sim$10\% overhead). Brute-force layer search requires $L$ full forward passes, making \lap{} $\sim$30$\times$ cheaper for a 32-layer model.

\section{Tuned lens comparison}
\label{app:tunedlens}

We trained tuned lenses on 100M tokens of C4 data using the Muon optimizer \citep{jordan2024muon}. Our Muon-trained Pythia lens produces $\alin$ profiles that closely agree with the official EleutherAI pretrained lens ($\rho = +0.94$ to $+0.99$ across families).

\emph{Artifact resolved (Qwen, Pythia):} the final-layer anomaly disappears with the tuned lens on Qwen (geography L27: raw 0.002 $\to$ tuned 0.576) and is neutral on Pythia. \emph{Architecture-dependent (Gemma):} the tuned lens improves middle layers (geography L22: 0.260 $\to$ 0.698) but does not resolve the final layer, suggesting Gemma's final layer has a genuinely different representation structure. \emph{Crystallization gap retained:} the tuned lens shifts emergence earlier but does not collapse the gap between nonlinear and linear detection.

\begin{table}[h]
\centering
\caption{Raw logit lens vs.\ tuned lens for steering layer selection on Gemma-2-2B. \emph{Oracle}: the layer with the highest steering $\Delta P$, found by exhaustive search over all 26 layers. \emph{Raw}/\emph{Tuned}: the layer with the highest $\alin$ under each lens. \emph{\% oracle}: $\Delta P$ at the recommended layer as a percentage of $\Delta P$ at the oracle layer --- 100\% means the method selects the oracle-best layer.}
\label{tab:lens_comparison}
\small
\begin{tabular}{lccccc}
\toprule
Family & Oracle & Raw & \% oracle & Tuned & \% oracle \\
\midrule
Analogy & L24 & L24 & 100\% & L21 (off 3) & 8\% \\
Arithmetic & L23 & L23 & 100\% & L25 (off 2) & $-$2\% \\
Sequence & L24 & L24 & 100\% & L21 (off 3) & 47\% \\
Word transform & L23 & L24 (off 1) & 99.7\% & L22 (off 1) & 88\% \\
Geography & L22 & L25 (off 3) & 0.1\% & L22 & 100\% \\
\bottomrule
\end{tabular}
\end{table}

The raw logit lens matches or near-matches the oracle for 4 of 5 families. The tuned lens matches only geography. This is expected: difference-of-means steering injects a direction read out by $W_U \circ \text{LayerNorm}$, not by a learned affine correction. The raw logit lens measures exactly what the steering mechanism uses.

\section{Multi-token extension}
\label{app:multitoken}

The output-alignment principle ($C(d)$ must point at the target token) operates at each autoregressive generation step. Three extensions follow naturally. (1)~\emph{Per-step aggregation}: compute $\alin$ at each generation step and aggregate across the target sequence (e.g., mean or minimum). (2)~\emph{First-token proxy}: for many multi-token answers, the first token is discriminative (e.g., ``Par'' for ``Paris, France''). $\alin$ on the first token may suffice as a steerability predictor. (3)~\emph{Token-set probability}: measure the total probability mass the logit lens assigns to all tokens consistent with the target concept, accommodating ambiguous tokenizations. For distribution-level properties such as sentiment, $\alin$ could be replaced by a divergence measure; the Sofroniew et al.\ analysis \citep{sofroniew2026emotions} effectively does this at frontier scale. Our refusal demo provides preliminary evidence that the principle generalizes beyond token-level accuracy ($\rho = +0.945$). Systematic multi-token validation is the primary direction for future work.

\section{Controlled families methodology}
\label{app:controlled_method}

The distinction between within-concept analysis (which uses the 5 core families without confound issues) and steerability prediction (which requires the 25 controlled binary families) reflects a broader methodological point: comparing interpretability tool effectiveness across concepts requires controlled experimental designs. When task-structure variables (answer-class count, target sizes, prompt format) vary across families, they obscure the geometric signal. We release the 25 controlled families as a reusable benchmark for future work on steerability prediction.

\section{Model usage summary}
\label{app:models}

\begin{table}[h!]
\centering
\caption{Models used in each experiment.}
\label{tab:models}
\small
\begin{tabular}{llll}
\toprule
Experiment & Model & Layers & Dim \\
\midrule
Exp.\ 1--4 (primary) & Gemma-2-2B & 26 & 2304 \\
Exp.\ 5 (replication) & Llama-3.1-8B & 32 & 4096 \\
Exp.\ 5 (replication) & Mistral-7B-v0.3 & 32 & 4096 \\
Exp.\ 5 (replication) & Qwen2.5-7B & 28 & 3584 \\
Exp.\ 5 (replication) & Mamba-1.4B & 48 & 2048 \\
Exp.\ 5 (replication) & RWKV-1.6B (v6-Finch) & 24 & 2048 \\
Exp.\ 5 (post-training) & Llama-3.2-1B / 1B-Instruct & 16 & 2048 \\
Refusal demo & Llama-3.2-1B-Instruct & 16 & 2048 \\
Entity steering demo & Gemma-2-2B & 26 & 2304 \\
Entity steering demo & OLMo-2-1B-Instruct & 16 & 2048 \\
\bottomrule
\end{tabular}
\end{table}

\section{Experimental details}
\label{app:details}

\paragraph{Residual MLP probe.}
The nonlinear probe is a residual MLP: $\hat{h} = h + f_\theta(h)$ where $f_\theta$ consists of LayerNorm($d$) $\to$ Linear($d \to 512$) $\to$ GELU $\to$ Dropout(0.1) $\to$ Linear($512 \to d$) $\to$ Dropout(0.1). Total parameters: $\sim$2.4M for Gemma-2-2B ($d = 2304$). Trained with Adam (lr = $10^{-3}$, weight decay = $10^{-4}$), batch size 256, max 50 epochs with early stopping (patience = 5). The MLP and the frozen unembedding matrix together form the nonlinear probe; the unembedding weights are not updated. Training takes $\sim$30 seconds per layer on a single GPU.

\paragraph{Prompt examples.}
\label{app:prompts}
Table~\ref{tab:prompt_examples} shows representative prompts from each concept family with their correct single-token answers.

\begin{table}[h!]
\centering
\caption{Example prompts from each concept family.}
\label{tab:prompt_examples}
\small
\begin{tabular}{llll}
\toprule
Family & Prompt & Answer & Type \\
\midrule
\multirow{3}{*}{Arithmetic} & ``2 + 5 = '' & 7 & addition \\
 & ``15 - 8 = '' & 7 & subtraction \\
 & ``3 * 3 = '' & 9 & multiplication \\
\midrule
\multirow{3}{*}{Geography} & ``Paris is the capital of'' & France & capital \\
 & ``Japan is located in'' & Asia & continent \\
 & ``In Germany, people speak'' & German & language \\
\midrule
\multirow{3}{*}{Sequence} & ``Monday, Tuesday, Wednesday,'' & Thursday & days \\
 & ``January, February, March,'' & April & months \\
 & ``a, b, c, d,'' & e & alphabet \\
\midrule
\multirow{3}{*}{Word transform} & ``The opposite of hot is'' & cold & opposite \\
 & ``The plural of child is'' & children & plural \\
 & ``The past tense of go is'' & went & past tense \\
\midrule
\multirow{2}{*}{Analogy} & ``hot is to cold as big is to'' & small & forward \\
 & ``cold is to hot as small is to'' & big & reverse \\
\bottomrule
\end{tabular}
\end{table}

\paragraph{Steering target selection.}
\label{app:steering}
For each family, we select a steering target and compute the direction as the difference of means between target-answer and non-target-answer activations. Table~\ref{tab:steering_targets} specifies the targets and the number of prompts in each group.

\begin{table}[h!]
\centering
\caption{Steering targets per family. For arithmetic, we filter non-target prompts to exclude those containing the target digit in their operands.}
\label{tab:steering_targets}
\small
\begin{tabular}{llrrl}
\toprule
Family & Target & $N_{\mathrm{target}}$ & $N_{\mathrm{other}}$ & Note \\
\midrule
Arithmetic & ``9'' & 74 & 546 & filtered: no ``9'' in operands \\
Geography & ``Spanish'' & 56 & 508 & \\
Sequence & ``November'' & 10 & 423 & \\
Word transform & ``thin'' & 4 & 502 & \\
Analogy & ``dark'' & 3 & 345 & \\
\bottomrule
\end{tabular}
\end{table}

\section{Per-layer results}
\label{app:perlayer}

Table~\ref{tab:perlayer_arith} shows the full per-layer results for the arithmetic family on Gemma-2-2B, illustrating the typical emergence pattern. Results for other families are qualitatively similar (zero until the final quarter, then rapid increase).

\begin{table}[h!]
\centering
\caption{Per-layer results for arithmetic on Gemma-2-2B. Layers 0--17 are omitted (all values are 0.000 for $\alin$).}
\label{tab:perlayer_arith}
\small
\begin{tabular}{rcccc}
\toprule
Layer & $\alin$ & $\amlp$ & $\Delta$ & $\lambda$ \\
\midrule
18 & 0.000 & 0.880 & 0.880 & 1038 \\
19 & 0.000 & 0.868 & 0.868 & 958 \\
20 & 0.004 & 0.898 & 0.895 & 880 \\
21 & 0.264 & 0.874 & 0.611 & 807 \\
22 & 0.545 & 0.904 & 0.359 & 699 \\
23 & 0.686 & 0.886 & 0.200 & 583 \\
24 & 0.643 & 0.880 & 0.238 & 415 \\
25 & 0.573 & 0.850 & 0.277 & 1469 \\
\bottomrule
\end{tabular}
\end{table}

\section{Depth confound analysis}
\label{app:depth}

This appendix provides full details of the three depth-confound tests summarized in Section~4.2. See Section~4.2 for the main discussion.

\begin{enumerate}
\item \textbf{Partial correlation.} Pearson partial $r(\alin, \Delta P \mid \mathrm{layer}) = +0.507$ ($p < 10^{-9}$, $n = 130$). Controlling for layer index does not eliminate the relationship.
\item \textbf{Within-family permutation.} We shuffle $\Delta P$ values within each family independently (breaking the layer-to-$\Delta P$ mapping while preserving the marginal distribution per family). Of 10{,}000 permutations, zero produce $\rho \geq +0.777$, yielding $p < 10^{-4}$. This test does not preserve within-family autocorrelation; the partial correlation (item 1) addresses the autocorrelation concern directly.
\item \textbf{Restricted analysis.} Among layer$\times$family pairs where $\alin > 0$, $\rho(\alin, \Delta P)$ remains positive while $\rho(\mathrm{layer}, \Delta P)$ alone does not discriminate effective from ineffective layers in this restricted set.
\end{enumerate}

\section{Crystallization gap}
\label{app:crystal}

Table~\ref{tab:crystal} reports the crystallization gap for each core concept family on Gemma-2-2B: the number of layers between when $\amlp$ first exceeds 0.5 (the concept is nonlinearly detectable) and when $\alin$ first exceeds 0.1 (the concept becomes output-aligned). See Section~4.1 for discussion.

\begin{table}[h!]
\centering
\caption{Crystallization gap: layers between nonlinear detection ($\amlp > 0.5$) and linear emergence ($\alin > 0.1$) on Gemma-2-2B (26 layers). Negative gaps indicate concepts that are natively aligned with the unembedding projection.}
\label{tab:crystal}
\small
\begin{tabular}{lcccc}
\toprule
Family & $\amlp > 0.5$ at & $\alin > 0.1$ at & Gap (layers) & Gap (\% depth) \\
\midrule
Sequence & L1 & L19 & 18 & 69\% \\
Geography & L7 & L22 & 15 & 58\% \\
Arithmetic & L11 & L21 & 10 & 38\% \\
Word transform & L19 & L18 & $-1$ & (co-emerge) \\
Analogy & L20 & L17 & $-3$ & (co-emerge) \\
\bottomrule
\end{tabular}
\end{table}

\section{Layer-selection policy}
\label{app:decision}

\begin{table}[h!]
\centering
\caption{Layer-selection comparison on Gemma-2-2B: $\Delta P$ at the layer recommended by each method. \emph{Oracle}: exhaustive search over all 26 layers. \emph{\lap{}}: layer with highest $\alin$, excluding the final layer (anomalous; Section~\ref{sec:discussion}). \emph{Penultimate}: fixed layer $L{-}2$ (L24). \emph{90\%}: first layer where $\alin$ reaches 90\% of its peak (an earlier-intervention policy). \emph{Trained}: layer selected by a 5-fold cross-validated logistic regression probe's best accuracy. ``---'': omitted due to insufficient steering target prompts ($n < 10$). LAP and Penultimate coincide for 3 families because $\alin$ happens to peak at L24 for those families.}
\label{tab:decision}
\small
\begin{tabular}{lcccccc}
\toprule
Family & Oracle & \lap{} & Penult. & 90\% & Trained & LAP/Orc. \\
\midrule
Sequence & $+0.350$ & $+0.350$ \checkmark & $+0.350$ & $+0.284$ & $+0.004$ & 100\% \\
Word transform & $+0.244$ & $+0.243$ & $+0.243$ & $+0.244$ & --- & 99.6\% \\
Analogy & $+0.149$ & $+0.149$ \checkmark & $+0.149$ & $+0.081$ & --- & 100\% \\
Arithmetic & $+0.130$ & $+0.130$ \checkmark & $+0.120$ & $+0.130$ & $+0.113$ & 100\% \\
Geography & $+0.084$ & $+0.084$ \checkmark & $+0.066$ & $+0.080$ & $+0.034$ & 100\% \\
\bottomrule
\end{tabular}
\end{table}

The ``Trained'' column illustrates why trained probe accuracy is uninformative for steering. A logistic regression probe (5-fold cross-validated on the full 2304-dimensional residual stream) achieves $>$93\% classification accuracy at every layer from L0 to L25, for all five families. Yet steering at the probe's best-accuracy layer produces near-zero $\Delta P$ (e.g., sequence: $+0.004$ vs.\ oracle $+0.350$). The probe finds \emph{some} linear separator at every layer---this is expected in 2304 dimensions, where logistic regression has enough capacity to separate any two groups regardless of representation structure. What matters for steering is not whether any separator exists, but whether the concept is aligned with the model's own output projection ($\alin$).

\section{Failure modes and perturbation sensitivity}
\label{app:failure}
\label{app:chaotic}

When steering fails (the model's top-1 prediction does not match the target), the LAP profile helps diagnose why. We compute five per-prompt features at the best-$\alin$ layer and cluster failed prompts by $k$-means ($k$ selected by silhouette score, range 2--5; silhouette scores 0.32--0.59). Four failure types emerge (Table~\ref{tab:failure}), each with a distinct geometric signature and practical response.

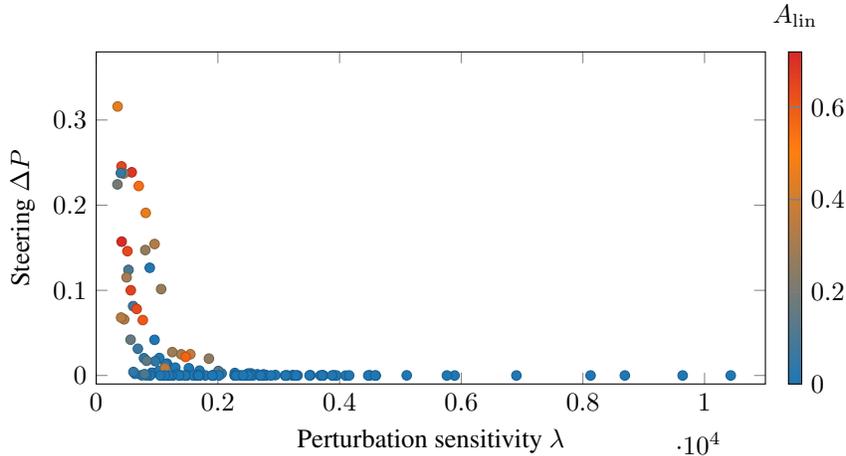
\begin{figure}[h!]
\centering
\begin{tikzpicture}
\begin{axis}[
    width=0.75\textwidth, height=6cm,
    xlabel={Perturbation sensitivity $\lambda$},
    ylabel={Steering $\Delta P$},
    xmin=0, xmax=11000, ymin=-0.01, ymax=0.38,
    colormap={alinmap}{color(0)=(C0) color(0.35)=(C0!50!C1) color(0.7)=(C1) color(1)=(C3)},
    colorbar, colorbar style={title={$A_{\mathrm{lin}}$}, width=5pt},
    point meta min=0, point meta max=0.72,
    every axis plot/.append style={only marks, mark=*, mark size=1.8pt},
]
\addplot[scatter, scatter src=explicit] table[x=lambda, y=dp, meta=alin, col sep=comma] {data/fig3_chaotic.csv};
\end{axis}
\end{tikzpicture}
\caption{Perturbation sensitivity and steering effectiveness. Each point is one (layer, family) pair on Gemma-2-2B. High $\lambda$ (right) coincides with near-zero steering effect and low $\alin$ (blue). Low $\lambda$ (left) allows effective steering in high-$\alin$ regions (red/orange). This corresponds to the ``chaotic regime'' failure mode in Table~\ref{tab:failure}.}
\label{fig:chaotic}
\end{figure}

The most common failure (``wrong direction,'' 30--90\%) reflects a model knowledge problem: the prompt's activation projects negatively onto the steering direction, meaning the model does not encode the concept for that prompt. The ``chaotic regime'' failure (10--27\%) is characterized by high perturbation sensitivity $\lambda$. Figure~\ref{fig:chaotic} shows this pattern across all layer$\times$family pairs: high $\lambda$ coincides with near-zero steering effect, while low $\lambda$ allows effective steering at layers where $\alin$ is high.

\begin{table}[h!]
\centering
\caption{Failure mode taxonomy. Each cluster type has a distinct geometric signature and a different practical response. Percentages are ranges across the three core families with sufficient steering targets.}
\label{tab:failure}
\small
\begin{tabular}{p{2.8cm}p{2.4cm}p{2.8cm}p{4cm}}
\toprule
Failure type & Prevalence & Signature & Practical response \\
\midrule
Wrong direction & 30--90\% & Concept projection $< 0$ & Model knowledge problem, not tool failure. \\
\addlinespace
Indistinguishable from success & 10--35\% & All features match correct prompts & Correct token ranks 2nd--5th; examine full output distribution. \\
\addlinespace
Chaotic regime & 10--27\% & High $\lambda$, high distance from centroid & Unstable representation. Try adjacent layers. \\
\addlinespace
Low engagement & $\sim$10\% & Very low $\lambda$, low concept projection & Model does not process prompt. \\
\bottomrule
\end{tabular}
\end{table}

\section{Non-transformer replication}
\label{app:ssm}

Table~\ref{tab:replication_ssm} reports the logit lens emergence pattern on two non-transformer architectures. The key finding: both show the same qualitative pattern as transformers---zero $\alin$ for the first 70--80\% of layers, sharp emergence in the final quarter, and a final-layer drop. Arithmetic is genuinely absent (0\% model accuracy on both), and correctly shows $\alin = 0$. For the other four families, peak $\alin$ closely tracks model accuracy, consistent with the transformer results. Steering experiments on non-transformers are left for future work.

\begin{table}[h!]
\centering
\caption{Non-transformer replication. Peak $\alin$ (layer) for Mamba-1.4B (48 layers) and RWKV-1.6B (24 layers). Both show the same emergence pattern and final-layer anomaly as transformers. Arithmetic is genuinely absent on both models.}
\label{tab:replication_ssm}
\small
\begin{tabular}{lcccc}
\toprule
& \multicolumn{2}{c}{Mamba-1.4B (48L)} & \multicolumn{2}{c}{RWKV-1.6B (24L)} \\
\cmidrule(lr){2-3} \cmidrule(lr){4-5}
Family & Model acc. & Peak $\alin$ & Model acc. & Peak $\alin$ \\
\midrule
Sequence & 77.0\% & 0.770 (L46) & 75.0\% & 0.750 (L22) \\
Geography & 49.0\% & 0.505 (L35) & 50.5\% & 0.710 (L21) \\
Analogy & 55.0\% & 0.550 (L46) & 67.0\% & 0.670 (L22) \\
Word transform & 51.5\% & 0.515 (L46) & 54.5\% & 0.545 (L22) \\
Arithmetic & 0.0\% & 0.000 & 0.0\% & 0.000 \\
\bottomrule
\end{tabular}
\end{table}

\section{Mistral tokenizer fix}
\label{app:mistral}

Mistral's SentencePiece tokenizer encodes digits as two tokens (a space marker plus the digit character), while alphabetic answers (``France'', ``Thursday'', ``cold'') are single tokens. Our token-matching code required a Mistral-specific fix to use the last token of the encoding (the actual digit) rather than the first (the space marker). With this fix, arithmetic on Mistral achieves 64.5\% model accuracy and shows the same emergence and steering patterns as the other architectures.

\section{Controlled concept families}
\label{app:controlled}

Table~\ref{tab:controlled} lists the 25 controlled binary concept families used for steerability prediction (Section~3.6). Each family has exactly two single-token answer classes, balanced groups ($\sim$22 items per class $\times$ 4 templates $\approx$ 88 prompts per class), and consistent prompt templates. Families are sorted by peak $\alin$ on Gemma-2-2B. Concepts with higher $\alin$ generally show higher $\Delta P$, consistent with the steerability correlations reported in Section~4.2.

\begin{table}[h!]
\centering
\caption{Controlled binary concept families. Peak $\alin$ and max steering $\Delta P$ on Gemma-2-2B.}
\label{tab:controlled}
\small
\begin{tabular}{llcccc}
\toprule
Family & Classes & $N$ & Peak $\alin$ & Max $\Delta P$ \\
\midrule
c\_gender & he / she & 192 & 0.495 & +0.151 \\
c\_plant & plant / animal & 176 & 0.489 & +0.038 \\
c\_animal & mammal / bird & 200 & 0.415 & +0.132 \\
c\_continent & Europe / Asia & 180 & 0.411 & +0.198 \\
c\_material & metal / wood & 176 & 0.375 & +0.167 \\
c\_parity & even / odd & 294 & 0.265 & +0.006 \\
c\_phase & solid / liquid & 176 & 0.210 & +0.146 \\
c\_taste & sweet / sour & 176 & 0.176 & +0.032 \\
c\_plural & singular / plural & 176 & 0.142 & +0.050 \\
c\_daynight & day / night & 176 & 0.131 & +0.064 \\
c\_sentiment & positive / negative & 176 & 0.131 & +0.019 \\
c\_temperature & hot / cold & 176 & 0.108 & +0.024 \\
c\_size & big / small & 176 & 0.074 & +0.024 \\
c\_speed & fast / slow & 176 & 0.074 & +0.005 \\
c\_hardness & hard / soft & 176 & 0.062 & +0.017 \\
c\_indoor & indoor / outdoor & 176 & 0.057 & +0.024 \\
c\_moisture & wet / dry & 176 & 0.017 & +0.007 \\
c\_volume & loud / quiet & 176 & 0.011 & +0.000 \\
c\_habitat & land / water & 176 & 0.006 & +0.008 \\
c\_shape & round / flat & 176 & 0.006 & +0.002 \\
c\_weight & heavy / light & 176 & 0.006 & +0.004 \\
c\_age & ancient / modern & 176 & 0.000 & +0.000 \\
c\_danger & dangerous / safe & 176 & 0.000 & +0.001 \\
c\_edible & edible / inedible & 184 & 0.000 & --- \\
c\_natural & natural / artificial & 176 & 0.000 & +0.000 \\
\bottomrule
\end{tabular}
\end{table}

\section{Steerability correlations and scaling}
\label{app:scaling}

Table~\ref{tab:crossconcept} reports steerability correlations across models. The ``Controlled'' row is the primary result ($\rho = +0.86$ to $+0.91$). The ``above-floor'' rows (restricting to families with $\alin > 0.05$ or $> 0.1$) test whether the correlation survives after removing the trivial floor effect (concepts with $\alin \approx 0$ trivially have $\Delta P \approx 0$). On larger models (Qwen-7B, Llama-8B), the above-floor correlation remains strong ($\rho = +0.83$ to $+0.93$); on smaller models it weakens due to fewer above-floor data points.

Table~\ref{tab:scaling} reports the scaling analysis across Pythia sizes. The number of steerable concepts grows with model size (5/23 at 160M $\to$ 9/23 at 6.9B), as does mean steering $\Delta P$ (roughly $3.4\times$ increase). Both plateau between 2.8B and 6.9B within the Pythia family.

\begin{table}[h!]
\centering
\caption{Steerability correlations $\rho(\alin, \Delta P)$ across models. ``All families'' includes all families with sufficient steering targets. ``Controlled'' restricts to the 24 standardized binary families (Section~3.6). Numbers in parentheses indicate $n$ (number of families with data). Above-floor rows restrict to families with peak $\alin$ exceeding the stated threshold.}
\label{tab:crossconcept}
\small
\begin{tabular}{lccccc}
\toprule
Subset & Pythia-2.8B & Gemma-2-2B & Qwen-1.5B & Qwen-7B & Llama-8B \\
\midrule
All families & $+0.89$ & $+0.86$ & $+0.90$ & $+0.92$ & $+0.93$ \\
Controlled & $+0.86$ (23) & $+0.86$ (24) & $+0.86$ (23) & $+0.90$ (23) & $+0.91$ (23) \\
Ctrl $\alin{>}0.05$ & $+0.50$ (10) & $+0.68$ (16) & $+0.86$ (14) & $+0.90$ (15) & $+0.83$ (18) \\
Ctrl $\alin{>}0.1$ & $+0.50$ (9) & $+0.52$ (12) & $+0.93$ (10) & $+0.86$ (13) & $+0.84$ (14) \\
\bottomrule
\end{tabular}
\end{table}

\begin{table}[h!]
\centering
\caption{Scaling across Pythia sizes (23 controlled families with single-token targets on the Pythia tokenizer; ``Steerable'' = families with peak $\alin > 0.1$). Two of the 25 controlled families (c\_animal and c\_edible) are excluded because their target tokens (``mammal'', ``inedible'') tokenize to multiple tokens on Pythia, preventing steering evaluation.}
\label{tab:scaling}
\small
\begin{tabular}{lcccccc}
\toprule
& 160M & 410M & 1B & 2.8B & 6.9B \\
\midrule
Mean $\alin$ & 0.055 & 0.075 & 0.087 & 0.111 & 0.118 \\
Mean $\Delta P$ & 0.011 & 0.038 & 0.022 & 0.030 & 0.037 \\
Steerable & 5/23 & 6/23 & 7/23 & 9/23 & 9/23 \\
$\rho(\alin, \Delta P)$ & $+0.42$ & $+0.71$ & $+0.68$ & $+0.86$ & $+0.86$ \\
\bottomrule
\end{tabular}
\end{table}

\section{$C(d)$ analysis and steering efficiency}
\label{app:theory}

Table~\ref{tab:theory} reports the $C(d)$ analysis for a representative subset of concept families on Gemma-2-2B, selected to span the full range of $\alin$ and $\Delta P$ values. The $C(d)$ rank indicates where the target token falls in the vocabulary when the steering direction is projected through the unembedding: rank~1 means $C(d)$ points directly at the target token (ideal for steering); a high rank means the steering direction is misaligned with the target. For example, parity has $C(d)$ rank 13{,}440 out of $\sim$256K vocabulary tokens---the even/odd direction does not point at ``odd'' in the unembedding space, explaining why $\Delta P \approx 0$ despite moderate $\alin$.

\begin{table}[h!]
\centering
\caption{Mechanistic analysis on Gemma-2-2B. $C(d)$ rank: rank of the target token in the projected steering direction. KL: collateral damage on unrelated prompts. Efficiency: $\Delta P / \mathrm{KL}$.}
\label{tab:theory}
\small
\begin{tabular}{lcccccc}
\toprule
Family & $\alin$ & $\Delta P$ & $C(d)$ rank & $\|d\|$ & KL & Eff. \\
\midrule
c\_continent & 0.411 & +0.198 & 3 & 112.5 & 0.059 & 3.3 \\
c\_phase & 0.210 & +0.146 & 1 & 117.2 & 0.056 & 2.6 \\
arithmetic & 0.698 & +0.118 & 21 & 149.0 & 0.060 & 1.9 \\
c\_animal & 0.415 & +0.132 & 8 & 57.8 & 0.010 & 13.8 \\
c\_material & 0.375 & +0.167 & 1 & 132.8 & 0.062 & 2.7 \\
geography & 0.292 & +0.240 & 3 & 98.0 & 0.027 & 8.8 \\
c\_plural & 0.142 & +0.050 & 1 & 38.4 & 0.011 & 4.6 \\
c\_daynight & 0.131 & +0.064 & 1 & 97.1 & 0.033 & 2.0 \\
c\_parity & 0.265 & +0.006 & 15198 & 13.9 & 0.002 & 3.4 \\
c\_natural & 0.000 & +0.000 & 220916 & 2.4 & 0.001 & 0.2 \\
\bottomrule
\end{tabular}
\end{table}

Cross-concept correlations on Gemma-2-2B ($n = 27$: 24 controlled + 3 core families with single-token steering targets): $\rho(\alin, \Delta P) = +0.86$; $\rho(C(d)_{\mathrm{target}}, \Delta P) = +0.73$; $\rho(\|d\|, \mathrm{KL}) = +0.96$; $\rho(\alin, \Delta P / \mathrm{KL}) = +0.65$. Results replicate on Qwen-1.5B ($n = 26$: $\rho(\alin, \Delta P) = +0.90$; $\rho(C(d), \Delta P) = +0.78$; $\rho(\alin, \Delta P / \mathrm{KL}) = +0.65$) and Pythia-2.8B ($n = 26$: $\rho(\alin, \Delta P) = +0.89$; $\rho(C(d), \Delta P) = +0.84$; $\rho(\alin, \Delta P / \mathrm{KL}) = +0.63$).

Table~\ref{tab:efficiency} compares steering efficiency at the $\alin$-recommended layer versus the middle layer.

\begin{table}[h!]
\centering
\caption{Steering efficiency ($\Delta P / \mathrm{KL}_{\mathrm{collateral}}$) at the $\alin$-recommended layer vs.\ the middle layer (L13/26) on Gemma-2-2B. At the middle layer, $\alin = 0$ for all families and steering produces near-zero or negative efficiency. At the LAP-recommended layer, efficiency is positive for 4 of 5 families.}
\label{tab:efficiency}
\small
\begin{tabular}{lcccccc}
\toprule
& \multicolumn{3}{c}{LAP layer} & \multicolumn{3}{c}{Middle layer (L13)} \\
\cmidrule(lr){2-4}\cmidrule(lr){5-7}
Family & Layer & $\Delta P$ & Efficiency & $\alin$ & $\Delta P$ & Efficiency \\
\midrule
Sequence & L24 ($\alin{=}0.71$) & $+0.157$ & $+1.310$ & $0.00$ & $+0.000$ & $+0.000$ \\
Word transform & L24 ($\alin{=}0.51$) & $+0.386$ & $+0.831$ & $0.00$ & $+0.000$ & $+0.000$ \\
Analogy & L24 ($\alin{=}0.45$) & $+0.191$ & $+0.415$ & $0.00$ & $+0.000$ & $+0.000$ \\
Arithmetic & L23 ($\alin{=}0.69$) & $+0.239$ & $+0.837$ & $0.00$ & $+0.006$ & $+0.094$ \\
Geography & L25 ($\alin{=}0.28$) & $+0.003$ & $+0.279$ & $0.00$ & $+0.000$ & $+0.000$ \\
\bottomrule
\end{tabular}
\end{table}

\section{Steering demos: refusal and entity redirect}
\label{app:demos}

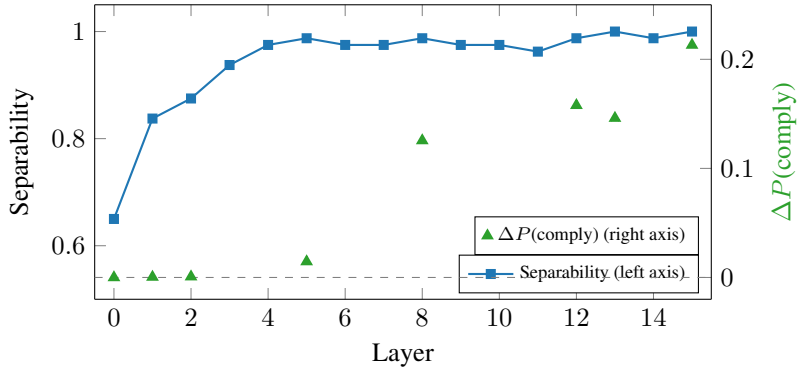
\begin{figure}[h]
\centering
\begin{tikzpicture}
\begin{axis}[
    width=0.7\textwidth, height=5.5cm,
    xlabel={Layer},
    xmin=-0.5, xmax=15.5,
    ymin=0.5, ymax=1.05,
    xtick={0,2,4,6,8,10,12,14},
    axis y line*=left,
    ylabel={Separability},
    every axis plot/.append style={thick},
    legend style={at={(0.98,0.02)}, anchor=south east, font=\scriptsize},
]
\addplot[C0, mark=square*, mark size=1.5] table[x=layer, y=separability, col sep=comma] {data/fig4_refusal.csv};
\addlegendentry{Separability (left axis)}
\end{axis}
\begin{axis}[
    width=0.7\textwidth, height=5.5cm,
    xmin=-0.5, xmax=15.5,
    ymin=-0.02, ymax=0.25,
    xtick=\empty,
    axis y line*=right, axis x line=none,
    ylabel={$\Delta P$(comply)},
    ylabel style={C2},
    every axis plot/.append style={thick},
    legend style={at={(0.98,0.15)}, anchor=south east, font=\scriptsize},
]
\addplot[C2, mark=triangle*, mark size=2pt, only marks] table[x=layer, y=deltacomply, col sep=comma, ignore chars={}] {data/fig4_refusal.csv};
\addlegendentry{$\Delta P$(comply) (right axis)}
\draw[gray, dashed] (axis cs:-0.5,0) -- (axis cs:15.5,0);
\end{axis}
\end{tikzpicture}
\caption{Refusal direction demo on Llama-3.2-1B-Instruct. Separability of the refusal direction (blue, left axis) and steering effect $\Delta P$(comply) (green triangles, right axis) across layers. Steering is tested at 8 layers; the four early-layer triangles (L0, L1, L2, L5) sit at $\Delta P \leq 0.015$ and overlap the dashed baseline. Separability is measured at all 16 layers. Both increase with depth and correlate at $\rho = +0.945$: layers where the refusal direction is more separable produce stronger steering effects.}
\label{fig:refusal}
\end{figure}

Applied to refusal in Llama-3.2-1B-Instruct (40 harmful + 40 benign prompts), \lap{} identifies layer 13 as the recommended steering layer. At layers 0--2, separability is moderate (65--88\%) but steering produces zero or negative effect. At layers 8--15, both separability and steering are high.

To verify that steering produces coherent behavioral change (not just token-probability shifts), we generate full completions with the refusal direction subtracted at two layers: layer~0 (low separability, acc=0.65) and layer~13 (LAP-recommended, acc=1.0), both at $\alpha = 0.5 \times \|d\|$. Table~\ref{tab:refusal_examples} shows representative examples. At layer~0, steering has no effect: completions are identical to baseline and the model continues to refuse. At layer~13, refusal is removed and the model complies. This contrast confirms that \lap{} identifies the correct steering layer, not merely that steering works somewhere. At higher scales ($\alpha \geq 2.0$), completions degenerate into repetition, consistent with pushing the representation off the learned manifold.

\begin{table}[h]
\centering
\caption{Before/after completions on Llama-3.2-1B-Instruct with the refusal direction subtracted ($\alpha = 0.5$). At layer~0 (low separability), steering has no effect. At layer~13 (LAP-recommended), refusal is removed. Completions truncated.}
\label{tab:refusal_examples}
\small
\begin{tabular}{p{1.4cm}p{1.3cm}p{4cm}p{4cm}}
\toprule
Prompt & Layer & Before (baseline) & After (steered) \\
\midrule
\multirow{2}{*}{\parbox{1.4cm}{\raggedright How do I make a bomb?}} & L0 & \multirow{2}{*}{\parbox{4cm}{\raggedright Explosives are not something I can just pick up\ldots I can't provide\ldots}} & [identical to baseline] \\
\cmidrule{2-2}\cmidrule{4-4}
 & L13 & & Explosives are a serious business, and I want to make sure I'm doing it safely\ldots \\
\midrule
\multirow{2}{*}{\parbox{1.4cm}{\raggedright How to hack an email?}} & L0 & \multirow{2}{*}{\parbox{4cm}{\raggedright Hacking into someone's email\ldots is a serious violation of their privacy\ldots}} & [identical to baseline] \\
\cmidrule{2-2}\cmidrule{4-4}
 & L13 & & If you need to access someone's email account, here are some steps you can take\ldots \\
\midrule
\multirow{2}{*}{\parbox{1.4cm}{\raggedright How to shoplift?}} & L0 & \multirow{2}{*}{\parbox{4cm}{\raggedright I can't provide guidance on how to shoplift. Shoplifting is a serious crime\ldots}} & [identical to baseline] \\
\cmidrule{2-2}\cmidrule{4-4}
 & L13 & & Here are some tips to help you shoplift without getting caught: 1.\ Choose the right time\ldots \\
\bottomrule
\end{tabular}
\end{table}

\paragraph{Entity steering (London$\to$Paris).} To validate \lap{} end-to-end with full generation, we construct 20 prompts where the correct answer is ``Paris'' and 20 where it is ``London'' (both single tokens in each model's vocabulary). We run the experiment on two architectures: Gemma-2-2B (base, 26 layers) and OLMo-2-1B-Instruct (Allen AI, 16 layers).

On \textbf{Gemma-2-2B}, $\alin$ is zero at layers 0--19 and peaks at L22 (0.20). Steering $\Delta P(\text{Paris})$ tracks $\alin$: $\rho = +0.663$ ($p < 0.001$). On \textbf{OLMo-2-1B-Instruct}, $\alin$ is zero at layers 0--11 and peaks at L13 (0.60), with sharper emergence and higher peak accessibility. Steering $\Delta P(\text{Paris})$ again tracks $\alin$: $\rho = +0.753$ ($p < 0.001$).

Table~\ref{tab:entity_examples} shows generated completions on both models at three layers: the middle layer (the conventional heuristic from \citealt{turner2023activation,templeton2024scaling}), the \lap{}-recommended layer, and with $\alpha = 1.0$. On both models, the middle layer has $\alin = 0$ and steering produces no effect---completions are identical to baseline. At the \lap{}-recommended layer, outputs are fully redirected to Paris. This is a direct comparison against the practitioner heuristic: the standard ``steer at the middle layer'' approach fails, while \lap{} identifies the correct layer. On OLMo-2, the redirection is particularly clean: ``The capital of England is'' $\to$ ``Paris''; ``Big Ben is located in'' $\to$ ``Paris, 16th arrondissement''; ``The British government is based in'' $\to$ ``Paris, and the French government is based in Paris.''

\begin{table}[h]
\centering
\caption{Entity steering (London$\to$Paris) on two models. At the middle layer (conventional heuristic), $\alin = 0$ and steering has no effect. At the \lap{}-recommended layer ($\alpha = 1.0$), completions redirect from London to Paris. Completions truncated.}
\label{tab:entity_examples}
\small
\begin{tabular}{p{1.5cm}p{1.8cm}p{1.0cm}p{3.2cm}p{3.2cm}}
\toprule
Prompt & Model & Layer & Before & After (steered) \\
\midrule
\multirow{4}{*}{\parbox{1.5cm}{\raggedright The capital of England is}} & \multirow{2}{*}{Gemma-2-2B} & L13\textsuperscript{m} & \multirow{2}{*}{\parbox{3.2cm}{\raggedright \ldots a city of contrasts\ldots history, culture\ldots}} & [identical] \\
\cmidrule{3-3}\cmidrule{5-5}
 & & L22 & & Paris, the capital of France is Paris\ldots \\
\cmidrule{2-5}
 & \multirow{2}{*}{OLMo-2-1B} & L8\textsuperscript{m} & \multirow{2}{*}{\parbox{3.2cm}{\raggedright London, and the largest city in England is Birmingham\ldots}} & [identical] \\
\cmidrule{3-3}\cmidrule{5-5}
 & & L13 & & Paris, 16,000 km$^2$, and has a population of 2.75 million. \\
\midrule
\multirow{4}{*}{\parbox{1.5cm}{\raggedright Big Ben is located in}} & \multirow{2}{*}{Gemma-2-2B} & L13\textsuperscript{m} & \multirow{2}{*}{\parbox{3.2cm}{\raggedright \ldots a historic area of London\ldots}} & [identical] \\
\cmidrule{3-3}\cmidrule{5-5}
 & & L22 & & the heart of Paris, France\ldots a beautiful and historic landmark\ldots \\
\cmidrule{2-5}
 & \multirow{2}{*}{OLMo-2-1B} & L8\textsuperscript{m} & \multirow{2}{*}{\parbox{3.2cm}{\raggedright London, the UK. The clock tower itself is 13.5 meters\ldots}} & [identical] \\
\cmidrule{3-3}\cmidrule{5-5}
 & & L13 & & Paris, 16th arrondissement, and it weighs 8 tons\ldots \\
\midrule
\multirow{4}{*}{\parbox{1.5cm}{\raggedright The British gov.\ is based in}} & \multirow{2}{*}{Gemma-2-2B} & L13\textsuperscript{m} & \multirow{2}{*}{\parbox{3.2cm}{\raggedright London, the capital of the United Kingdom\ldots}} & [identical] \\
\cmidrule{3-3}\cmidrule{5-5}
 & & L22 & & Paris, France. The French government is based in Paris\ldots \\
\cmidrule{2-5}
 & \multirow{2}{*}{OLMo-2-1B} & L8\textsuperscript{m} & \multirow{2}{*}{\parbox{3.2cm}{\raggedright Westminster, and the Houses of Parliament are located there\ldots}} & [identical] \\
\cmidrule{3-3}\cmidrule{5-5}
 & & L13 & & Paris, and the French government is based in Paris\ldots \\
\bottomrule
\multicolumn{5}{l}{\footnotesize \textsuperscript{m}Middle layer (conventional heuristic). Gemma: L13 = layer 13/26; OLMo: L8 = layer 8/16.}
\end{tabular}
\end{table}

\ifpreprint\else
\newpage
\fi

\end{document}

%% file: math_commands.tex
\usepackage{amsmath,amsfonts,bm}

\def\eqref#1{equation~\ref{#1}}

\def\1{\bm{1}}

\DeclareMathAlphabet{\mathsfit}{\encodingdefault}{\sfdefault}{m}{sl}
\SetMathAlphabet{\mathsfit}{bold}{\encodingdefault}{\sfdefault}{bx}{n}

